\documentclass{ecai} 

\usepackage{latexsym}
\usepackage{amssymb}
\usepackage{amsmath}
\usepackage{amsthm}
\usepackage{booktabs}
\usepackage{enumitem}
\usepackage{graphicx}
\usepackage{multirow} 
\usepackage{color}

\newcommand{\BibTeX}{B\kern-.05em{\sc i\kern-.025em b}\kern-.08em\TeX}

\begin{document}


\begin{frontmatter}


\paperid{123} 


\title{A Self-Adaptive Frequency Domain Network for Continuous Intraoperative Hypotension Prediction}


\author[A]{\fnms{Xian}~\snm{Zeng}}
\author[A]{\fnms{Tianze}~\snm{Xu}}
\author[A]{\fnms{Kai}~\snm{Yang}}
\author[B]{\fnms{Jie}~\snm{Sun}}
\author[B]{\fnms{Youran}~\snm{Wang}}
\author[A]{\fnms{Jun}~\snm{Xu}}
\author[A]{\fnms{Mucheng}~\snm{Ren}\thanks{Corresponding Author. Email: renm@nuist.edu.cn.}}

\address[A]{Jiangsu Key Laboratory of Intelligent Medical Image Computing, School of Artificial Intelligence, Nanjing University of Information Science and Technology, China}
\address[B]{Department of Anesthesiology, Surgery and Pain Management, School of Medicine, Zhongda Hospital, Southeast University, China}


\begin{abstract}
Intraoperative hypotension (IOH) is strongly associated with postoperative complications, including postoperative delirium and increased mortality, making its early prediction crucial in perioperative care. While several artificial intelligence-based models have been developed to provide IOH warnings, existing methods face limitations in incorporating both time and frequency domain information, capturing short- and long-term dependencies, and handling noise sensitivity in biosignal data. To address these challenges, we propose a novel \textbf{S}elf-\textbf{A}daptive \textbf{F}requency \textbf{D}omain \textbf{Net}work (\textbf{SAFDNet}). Specifically, SAFDNet integrates an adaptive spectral block, which leverages Fourier analysis to extract frequency-domain features and employs self-adaptive thresholding to mitigate noise. Additionally, an interactive attention block is introduced to capture both long-term and short-term dependencies in the data. Extensive internal and external validations on two large-scale real-world datasets demonstrate that SAFDNet achieves up to 97.3\% AUROC in IOH early warning, outperforming state-of-the-art models. Furthermore, SAFDNet exhibits robust predictive performance and low sensitivity to noise, making it well-suited for practical clinical applications.
\end{abstract}

\end{frontmatter}


\section{Introduction}
Intraoperative hypotension (IOH) is a critical concern in perioperative care because of its strong association with severe postoperative complications, such as significantly increased mortality rates, myocardial injury, acute kidney injury, postoperative delirium, and other neurological deficits~\cite{devereaux2012association,van2016association,wesselink2018intraoperative,wachtendorf2022association}. The likelihood of severe complications rises with the length of time hypotension persists, but it can start to develop in just a few minutes~\cite{walsh2014relationship}. Advance warning of IOH events during the perioperative period is a highly effective approach to prevent postoperative complications, even if the warning comes only 10 to 15 minutes ahead~\cite{sessler2019perioperative}. It allows clinicians to take proactive measures respond to hypotension, shifting from the traditional reactive management of hypotensive episodes to addressing them before adverse outcomes occurs. 
\begin{figure}[th]
\centering
    \includegraphics[width=\linewidth]{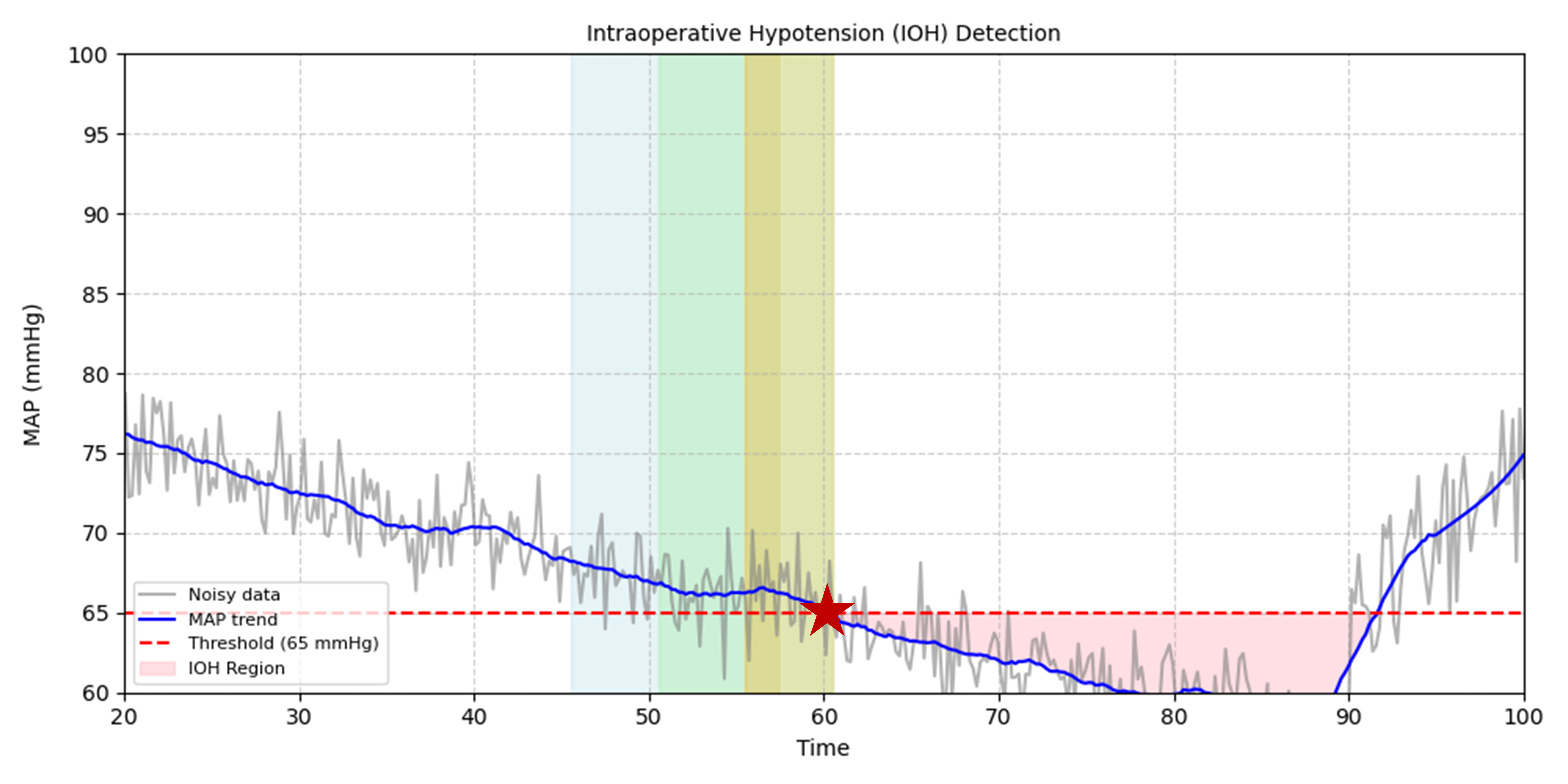}
\caption{The illustration of IOH prediction with 3, 5, 10, and 15-minute prediction windows.} \label{fig:intro}
\end{figure} 
Biosignal monitoring devices continuously collect waveform data, such as arterial blood pressure (ABP), electrocardiograms (ECG), carbon dioxide levels (CO2), and photoplethysmography (PPG), during surgery to monitor patient status and prevent adverse events~\cite{wijnberge2020effect,hosanee2020cuffless}. Figure~\ref{fig:intro} illustrates the IOH prediction process, where IOH is identified when the mean arterial pressure (MAP) falls below 65 mmHg for a sustained period. 

Artificial intelligence (AI)-based models have been developed to predict IOH using biosignal data~\cite{hatib2018machine,lee2021deep,lee2022intraoperative,lu2023composite,hwang2023intraoperative,kapral2024development,yang2024dynamic}, with some, such as the Hypotension Prediction Index (HPI)~\cite{hatib2018machine}, achieving commercial success.

While these models demonstrate potential, they face the following three critical limitations: 1) 
 \textbf{Underutilization of Frequency-Domain Information:} Most models focus on time-domain features, overlooking valuable frequency-domain insights. However, we hypothesize that frequency-domain information can reveal complementary patterns (e.g., spectral and periodic characteristics) that improve predictive performance~\cite{zhou2022fedformer,yi2024frequency}. 2) \textbf{Suboptimal Temporal Dependency Modeling:} Current models face trade-offs in temporal modeling: Transformers excel at capturing long-term dependencies but struggle with local patterns~\cite{kapral2024development,yang2024dynamic}, while CNNs are effective for short-term patterns but fail to model global dynamics~\cite{lee2021deep,lu2023composite}. We hypothesize that integrating these approaches can better capture hierarchical temporal relationships, improving IOH prediction accuracy. 3) \textbf{Sensitivity to Noise:} Real-world biosignal data are often noisy due to patient variability and device limitations. As shown in Figure~\ref{fig:intro}, noise and variability in biosignal data complicate trend analysis and prediction, and current models lack mechanisms to suppress noise while preserving critical signal information, limiting their robustness.

To address these challenges, we propose a novel framework, \textbf{S}elf-\textbf{A}daptive \textbf{F}requency \textbf{D}omain \textbf{Net}work (\textbf{SAFDNet}), for IOH prediction. SAFDNet predicts IOH events 3,5,10, and 15 minutes before their onset using biosignal data, such as ABP, ECG, CO2, and PPG waveforms. We evaluate SAFDNet on two large-scale datasets: VitalDB~\cite{lee2022vitaldb} and ZhongDa Vital (a private external validation dataset). SAFDNet achieves an AUROC of up to 97.3\% for IOH early warning, outperforming state-of-the-art models. Furthermore, it exhibits robust predictive performance and low sensitivity to noise, ensuring reliable operation in real-world clinical scenarios. These results underscore the practicality and effectiveness of SAFDNet in improving perioperative care.

Our contributions are summarized as follows:
\begin{itemize}
    \item We propose a \textbf{Self-Adaptive Filter Block} that leverages Fourier analysis to extract frequency-domain features while suppressing noise through a learnable, self-adaptive thresholding mechanism.
    \item We introduce a \textbf{Dual-Path Interactive Attention Block} that combines short-term and long-term temporal dependencies using a dual-path cross-attention mechanism between CNN and LSTM representations.
    \item We demonstrate the effectiveness of SAFDNet through extensive experiments and ablation studies on two real-world surgical datasets. SAFDNet achieves state-of-the-art performance with robust noise resilience, achieving an AUROC of up to 97.3\%.
\end{itemize}


\section{Related Works}

The prediction of IOH has been extensively studied using both time-domain models and frequency-domain forecasting techniques. Despite progress, existing works face challenges in effectively utilizing frequency-domain insights and robustly handling noise in biosignal data. Below, we review related works and highlight the gaps addressed by our proposed framework.

\subsection{Hypotension Prediction Models}
IOH prediction models rely heavily on biosignal waveforms and machine learning techniques. The Hypotension Prediction Index (HPI)~\cite{hatib2018machine,wijnberge2020effect} is a widely used algorithm that utilizes invasive arterial waveforms and additional features to predict IOH, setting a benchmark for clinical applications. Building on this, Lee et al.~\cite{lee2021deep} introduced a 1-D CNN model trained on 30-second waveform segments for predictions 5, 10, and 15 minutes in advance. Hwang et al.~\cite{hwang2023intraoperative} developed an interpretable model that predicts IOH 10 minutes ahead using 90-second arterial blood pressure (ABP) recordings, while Yang et al.~\cite{yang2024dynamic} combined CNNs and Transformers to capture both local and global dependencies in multichannel biosignals.

In addition to waveform-driven approaches, some methods incorporate clinical features. Lu et al.~\cite{lu2023composite} proposed a composite multi-attention framework that integrates demographic characteristics and low-sampling-rate waveform data for user-definable IOH event predictions. Similarly, Kapral et al.~\cite{kapral2024development} applied a Temporal Fusion Transformer to predict intraoperative blood pressure trajectories 7 minutes ahead using low resolution vital sign data.

While these methods have advanced IOH prediction, they primarily rely on time-domain features and focus on either short-term (CNN) or long-term (Transformer) dependencies. They often neglect frequency-domain characteristics, which are valuable for understanding periodic patterns in biosignals. Additionally, existing models lack robust mechanisms for mitigating noise and redundancy in real-world clinical data. To address these gaps, our proposed \textbf{SAFDNet} integrates Fourier analysis to leverage frequency-domain information and applies self-adaptive thresholding to suppress noise, ensuring improved robustness in noisy clinical environments.

\subsection{Forecasting in the Frequency Domain}
Recent advancements in time-series forecasting have demonstrated the benefits of frequency-domain techniques for capturing periodic patterns and long-term dependencies. For example, SFM~\cite{zhang2017stock} decomposes hidden states into multiple frequency components to model specific periodicities. FEDformer~\cite{zhou2022fedformer} enhances attention mechanisms by incorporating frequency-domain information, where weights are computed using the Discrete Fourier Transform (DFT). FreTS~\cite{yi2024frequency} transforms signals into the frequency domain using DFT and applies separate MLPs to process real and imaginary components. Similarly, Autoformer~\cite{wu2021autoformer} replaces self-attention with auto-correlation mechanisms implemented via Fast Fourier Transforms (FFT), and APDNet~\cite{zhuang2024rethinking} analyzes periodicity by decoupling real and imaginary parts of signals in the complex frequency domain.

While these methods improve general time-series forecasting, their application to clinical biosignals remains limited. Specifically, they do not address the challenges of noise and variability inherent in biosignal data from real-world clinical settings. In contrast, our proposed \textbf{SAFDNet} builds on these frequency-domain advancements by introducing a \textbf{Self-Adaptive Filter Block}, which extracts frequency-domain features and applies adaptive thresholding to suppress noise while preserving critical signal information. This mechanism ensures robust and accurate predictions, even with noisy and redundant biosignal data.

\section{Methodology}
\begin{figure*}[t]
\centering
    \includegraphics[width=\textwidth]{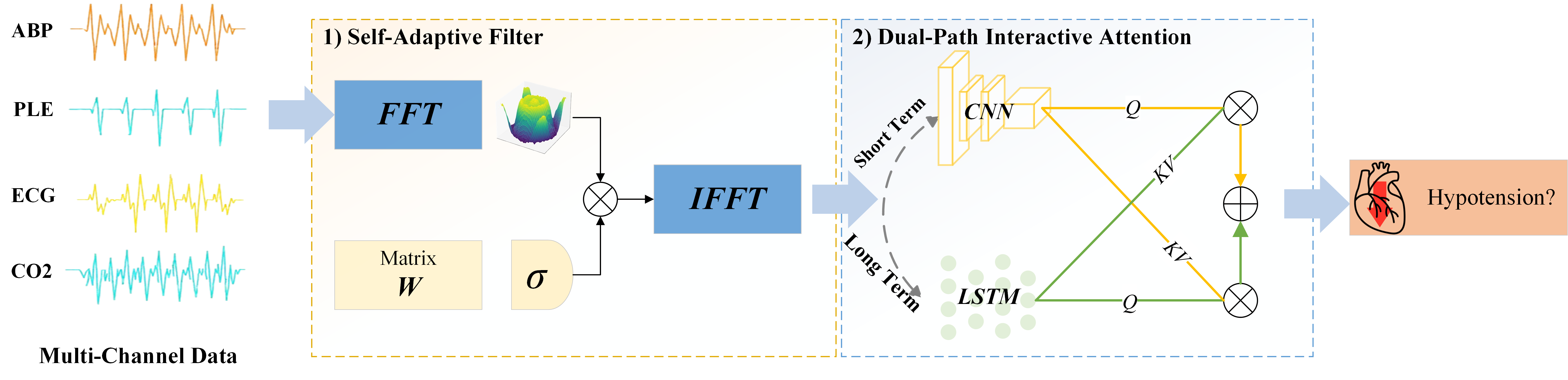}
\caption{The flowchart of proposed SAFDNet framework.} \label{fig:flow}
\end{figure*}

To address the challenges of noise, the underutilization of frequency-domain information, and the need for robust temporal dependency modeling in physiological waveform data, this study introduces \textbf{SAFDNet}, a two-stage framework for multi-channel waveform data. SAFDNet consists of two key components: (1) a \textbf{Self-Adaptive Filter Block}, which dynamically identifies and suppresses noisy or irrelevant frequency components while retaining task-relevant features, and (2) a \textbf{Dual-Path Interactive Attention Block}, which captures both short-term and long-term temporal dependencies through a dual-path attention mechanism. A schematic representation of the workflow is shown in Figure~\ref{fig:flow}.

\subsection{Self-Adaptive Filter Block}

Physiological waveforms, such as ABP, PPG, ECG, and CO2, are often contaminated by noise due to sensor inaccuracies, patient movement, and environmental interference. Additionally, not all frequency components contribute equally to IOH prediction tasks. Existing fixed filters, based on static cutoff frequencies, are insufficient for such dynamic signals, as they may discard valuable information or retain irrelevant noise. To address these issues, we propose the \textbf{Self-Adaptive Filter Block}.

The filtering process begins by transforming the multi-channel time-domain waveform \( x(t) \in \mathbb{R}^{C \times T} \), where \( C \) is the number of channels and \( T \) is the number of time samples, into the frequency domain using the \textit{Fast Fourier Transform (FFT)}. This results in the frequency-domain representation \( X(f) = \text{FFT}(x(t)) \), where \( X(f) \in \mathbb{C}^{C \times F} \), and \( F \) denotes the number of frequency bins. 

To enable dynamic filtering, we introduce a trainable parameter matrix \( W(f) \in \mathbb{R}^{C \times F} \), which learns the task-specific relevance of each frequency component. To map \( W(f) \) to the range \([0, 1]\), we apply the sigmoid activation function, where $\sigma$\( (W(f)) \) serves as the importance mask:
\begin{align}
\sigma(W(f)) = \frac{1}{1 + e^{-W(f)}}.
\end{align}
This produces an \textbf{importance mask}, where higher values emphasize relevant frequency components while suppressing less informative ones. The filtered frequency-domain signal is computed as:
\begin{align}
X'(f) = X(f) \odot \sigma(W(f)),
\end{align}
where \( \odot \) denotes element-wise multiplication. Unlike static filters, this adaptive mechanism allows the model to dynamically learn and adjust the filtering process based on the specific characteristics of the input data and prediction task, resulting in more robust feature extraction.

Finally, the filtered signal is transformed back into the time domain using the \textit{Inverse Fast Fourier Transform (IFFT)}, yielding the enhanced time-domain waveform:
\begin{align}
\hat{x}(t) = \text{IFFT}\big(\text{FFT}(x(t)) \odot \sigma(W(f))\big),
\end{align}
where \( \hat{x}(t) \in \mathbb{R}^{C \times T} \) represents the filtered signal. This self-adaptive filtering approach effectively enhances the quality of extracted features, providing a strong foundation for downstream temporal modeling.

\subsection{Dual-Path Interactive Attention Block}

To model both short-term and long-term temporal dependencies in the filtered time-domain data, we employ a dual-path architecture combining a \textbf{Convolutional Neural Network (CNN)} and a \textbf{Long Short-Term Memory (LSTM)} network. The CNN captures local temporal patterns (short-term dependencies), while the LSTM models global temporal relationships (long-term dependencies). These complementary representations are fused using a \textbf{dual-path cross-attention mechanism}, enabling rich interaction between short-term and long-term features for a comprehensive temporal understanding.

Let the outputs of the CNN and LSTM be represented as \( H_{\text{CNN}} \in \mathbb{R}^{T_s \times d_s} \) and \( H_{\text{LSTM}} \in \mathbb{R}^{T_l \times d_l} \), respectively, where \( T_s \) and \( T_l \) denote the sequence lengths, and \( d_s \), \( d_l \) are the feature dimensions. These outputs are integrated using two cross-attention mechanisms: \textbf{Short-to-Long Cross Attention} and \textbf{Long-to-Short Cross Attention}.

\paragraph{Short-to-Long Cross Attention}

In the short-to-long cross-attention mechanism, \( H_{\text{CNN}} \) serves as the query (\( Q \)), while \( H_{\text{LSTM}} \) acts as the key (\( K \)) and value (\( V \)). The attention is computed as:
\begin{align}
\alpha_{\text{s2l}} &= \text{softmax}\left( \frac{Q_{\text{CNN}} K_{\text{LSTM}}^\top}{\sqrt{d_k}} \right) V_{\text{LSTM}}, 
\end{align}

where \( Q_{\text{CNN}} = H_{\text{CNN}} W_q \), \( K_{\text{LSTM}} = H_{\text{LSTM}} W_k \), and \( V_{\text{LSTM}} = H_{\text{LSTM}} W_v \), \( W_q \in \mathbb{R}^{d_s \times d_k} \), \( W_k \in \mathbb{R}^{d_l \times d_k} \), and \( W_v \in \mathbb{R}^{d_l \times d_v} \), with \( d_k \) and \( d_v \) denoting the dimensionalities of the key and value spaces.

\paragraph{Long-to-Short Cross Attention}

Conversely, in the long-to-short cross-attention mechanism, \( H_{\text{LSTM}} \) serves as the query (\( Q \)), while \( H_{\text{CNN}} \) acts as the key (\( K \)) and value (\( V \)). The computation follows a similar formulation:
\begin{align}
\alpha_{\text{l2s}} &= \text{softmax}\left( \frac{Q_{\text{LSTM}} K_{\text{CNN}}^\top}{\sqrt{d_k}} \right) V_{\text{CNN}}, 
\end{align}
where \( Q_{\text{LSTM}} = H_{\text{LSTM}} W_q' \), \( K_{\text{CNN}} = H_{\text{CNN}} W_k' \), and \( V_{\text{CNN}} = H_{\text{CNN}} W_v' \), \( W_q' \in \mathbb{R}^{d_l \times d_k} \), \( W_k' \in \mathbb{R}^{d_s \times d_k} \), and \( W_v' \in \mathbb{R}^{d_s \times d_v} \) are learnable projection matrices.

\subsubsection{Feature Fusion and Prediction}

The outputs of the two cross-attention mechanisms, \( \alpha_{\text{s2l}} \in \mathbb{R}^{T_s \times d_v} \) and \( \alpha_{\text{l2s}} \in \mathbb{R}^{T_l \times d_v} \), are flattened into one-dimensional feature vectors, \( F_{\text{s2l}} \in \mathbb{R}^{d_f} \) and \( F_{\text{l2s}} \in \mathbb{R}^{d_f} \), respectively. These vectors are concatenated to form a unified representation:
\begin{align}
F_{\text{fused}} = [F_{\text{s2l}}; F_{\text{l2s}}],
\end{align}
where \( [\cdot; \cdot] \) denotes vector concatenation.

The fused feature vector \( F_{\text{fused}} \) is passed through a classification layer comprising a linear transformation, dropout for regularization, and a sigmoid activation function to produce the final prediction. By integrating short-term and long-term dependencies through the cross-attention mechanism, the proposed architecture achieves robust and accurate predictions for the target task.

\section{Experiment}
\subsection{Datasets}
We utilized two real-world datasets in our experiments: the publicly available VitalDB dataset~\cite{lee2022vitaldb}, designed for machine learning research on vital signs, and the private ZhongdaVital dataset collected from Zhongda Hospital affiliated with Southeast University.

VitalDB was developed by Seoul National University Hospital and contains comprehensive intraoperative data from 6,388 patients who underwent non-cardiac surgeries between June 2016 and August 2017. The dataset includes 486,451 waveform traces across 196 intraoperative monitoring parameters, 73 perioperative clinical parameters, and 34 time-series laboratory outcome parameters, providing a rich source of biosignals for analysis.

ZhongdaVital was collected from 465 patients who underwent non-cardiac surgeries between March 2024 and September 2024. This dataset includes vital sign data recorded by intraoperative monitoring devices, focusing on ABP, ECG, PPG, and CO2 waveforms. To ensure consistency in analysis, only patients with complete recordings of all four waveform types were included in this study.

The goal of both datasets was to develop and evaluate a model for predicting intraoperative hypotension (IOH). Table~\ref{tab:patient-characteristics} summarizes the demographic and clinical characteristics of the patients included in the experiments.

\begin{table}[!t]
\centering
\caption{Patient characteristics for internal (VitalDB) and external (ZhongDaVital) validation datasets, where * stands for American Society of Anesthesiologists.}
\resizebox{\linewidth}{!}{%
\begin{tabular}{lcc}
\toprule[2pt]
\textbf{Characteristics} & \textbf{VitalDB (n=3,369)} & \textbf{ZhongDaVital (n=465)} \\ \midrule
Age (years)              & 61.0 [50.0, 70.0]         & 64.0 [55.0, 72.0]            \\
Sex (male/female)        & 1,865 / 1,504             & 244 / 221                    \\
Weight (kg)              & 60.3 [53.2, 68.5]         & 65.0 [58.0, 72.0]            \\
Height (cm)              & 162.7 [156.4, 169.0]      & 165.0 [160.0, 170.0]         \\ \midrule
\textbf{ASA* Grade}       &                           &                               \\
1                        & 728                       & 10                            \\
2                        & 2,053                     & 324                           \\
3                        & 471                       & 116                           \\
4                        & 28                        & 15                            \\
5                        & 11                        & 0                             \\
Undefined                & 78                        & 0                             \\ \midrule
\textbf{Type of Anesthesia} &                       &                               \\
General                  & 3,350                     & 463                           \\
Spinal                   & 15                        & 1                             \\
Sedation/Analgesia       & 4                         & 1                             \\ \bottomrule[2pt]
\end{tabular}%
}

\label{tab:patient-characteristics}
\end{table}

\subsection{Data Processing}
The sampling rates of the dynamic sequences in VitalDB and ZhongdaVital ranged from 500 Hz to 100 Hz. To ensure consistent alignment across both datasets, we resampled all signals to 100 Hz, with each data segment spanning 30 seconds, resulting in 3,000 data points per sample. A peak detection algorithm was applied to the raw waveform data to identify individual cardiac cycles, with parameters such as height, prominence, and distance customized for each signal type to achieve precise segmentation~\cite{lee2021deep}. 

Several preprocessing steps were implemented to ensure data quality:
\begin{itemize}
    \item Segments with abnormal cardiac rhythms (e.g., excessively slow, fast, or undetectable) were excluded to eliminate noise and artifacts.
    \item Segments containing abnormal values in any of the four signals, such as mean arterial pressure (MAP) $<$ 20 mmHg or MAP $>$ 160 mmHg, were also excluded.
\end{itemize}

The target outcomes (1 = hypotension, 0 = nonhypotension) were defined based on MAP, which was calculated using systolic and diastolic blood pressures. Following prior research~\cite{hatib2018machine,lee2021deep}, hypotension was defined as periods where MAP remained $<$ 65 mmHg for at least 1 minute, while nonhypotension was defined as periods where MAP remained $>$ 75 mmHg for at least 1 minute. Periods with MAP values between 65 and 75 mmHg were considered a "gray zone" and excluded from model training to avoid ambiguity. 

For training, 30-second input sequences were used to predict hypotensive events 3, 5, 10, and 15 minutes before their occurrence. To balance the dataset, up to two nonhypotensive samples were selected from nonhypotensive periods for each hypotensive event. This preprocessing ensured high-quality, balanced inputs for the prediction model.

\begin{table*}[!t]
\centering
\caption{AUROC performance comparison between our method and other benchmark models.}
\resizebox{\textwidth}{!}{%
\begin{tabular}{llcccc}
\toprule[2pt]
\textbf{Method Type} & \textbf{Model}  & \textbf{Prediction = 3 min}& \textbf{5 min} & \textbf{10 min} & \textbf{15 min} \\ \midrule
\multicolumn{6}{c}{\textbf{Only ABP}} \\ \midrule
\multirow{3}{*}{\textbf{Conventional}} & KNN & 0.915 (0.913--0.917) & 0.892 (0.890--0.894) & 0.859 (0.857--0.862) & 0.852 (0.848--0.856) \\
 & Random Forest & 0.898 (0.896--0.901) & 0.881 (0.879--0.884) & 0.866 (0.863--0.869) & 0.844 (0.842--0.846) \\
 & SVM & 0.904 (0.903--0.906) & 0.886 (0.883--0.889) & 0.861 (0.858--0.864) & 0.861 (0.858--0.864) \\ \midrule
\multirow{4}{*}{\textbf{Reconstruction-based}} & VAE & 0.934 (0.932--0.935) & 0.920 (0.918--0.921) & 0.901 (0.899--0.903) & 0.892 (0.889--0.894) \\
 & USAD & 0.938 (0.936--0.940) & 0.924 (0.921--0.927) & 0.907 (0.903--0.909) & 0.900 (0.897--0.902) \\
 & Anomaly Transformer & 0.675 (0.672--0.679) & 0.653 (0.650--0.656) & 0.672 (0.668--0.676) & 0.669 (0.666--0.672) \\
 & MEMTO & 0.915 (0.912--0.918) & 0.884 (0.883--0.885) & 0.868 (0.866--0.870) & 0.862 (0.859--0.864) \\ \midrule
\multirow{5}{*}{\textbf{Classification-based}} & DNN & 0.905 (0.904--0.907) & 0.879 (0.877--0.882) & 0.857 (0.855--0.860) & 0.852 (0.848--0.857) \\
 & LSTM & 0.929 (0.927--0.930) & 0.908 (0.906--0.910) & 0.889 (0.887--0.892) & 0.886 (0.883--0.889) \\
 & GDN & 0.949 (0.947--0.951) & 0.929 (0.928--0.930) & 0.898 (0.895--0.900) & 0.896 (0.894--0.892) \\
 & Informer & 0.946 (0.944--0.947) & 0.924 (0.923--0.926) & 0.892 (0.889--0.894) & 0.890 (0.888--0.892) \\
 & \textbf{SAFDNet (Ours)} & \textbf{0.968 (0.966--0.971)} & \textbf{0.949 (0.947--0.950)} & \textbf{0.928 (0.927--0.929)} & \textbf{0.915 (0.913--0.917)} \\ \midrule
\multicolumn{6}{c}{\textbf{Multi-Channel}} \\ \midrule
\multirow{3}{*}{\textbf{Conventional}} & KNN & 0.926 (0.925--0.928) & 0.898 (0.896--0.899) & 0.872 (0.869--0.874) & 0.871 (0.868--0.873) \\
 & Random Forest & 0.918 (0.916--0.920) & 0.910 (0.908--0.912) & 0.890 (0.889--0.892) & 0.876 (0.873--0.879) \\
 & SVM & 0.909 (0.906--0.911) & 0.900 (0.898--0.902) & 0.876 (0.873--0.878) & 0.867 (0.864--0.870) \\ \midrule
\multirow{4}{*}{\textbf{Reconstruction-based}} & VAE & 0.938 (0.935--0.940) & 0.930 (0.927--0.931) & 0.921 (0.920--0.923) & 0.913 (0.910--0.916) \\
 & USAD & 0.947 (0.946--0.948) & 0.935 (0.933--0.937) & 0.923 (0.921--0.925) & 0.916 (0.915--0.917) \\
 & Anomaly Transformer & 0.892 (0.890--0.894) & 0.856 (0.852--0.859) & 0.837 (0.835--0.839) & 0.798 (0.794--0.801) \\
 & MEMTO & 0.922 (0.919--0.925) & 0.904 (0.902--0.906) & 0.892 (0.890--0.893) & 0.892 (0.890--0.894) \\ \midrule
\multirow{5}{*}{\textbf{Classification-based}} & DNN & 0.922 (0.921--0.924) & 0.894 (0.892--0.896) & 0.867 (0.864--0.870) & 0.853 (0.851--0.856) \\
 & LSTM & 0.946 (0.944--0.948) & 0.912 (0.910--0.914) & 0.901 (0.899--0.903) & 0.903 (0.901--0.905) \\
 & GDN & 0.952 (0.950--0.953) & 0.934 (0.931--0.937) & 0.911 (0.910--0.913) & 0.911 (0.908--0.913) \\
 & Informer & 0.951 (0.950--0.952) & 0.934 (0.933--0.935) & 0.911 (0.908--0.913) & 0.906 (0.904--0.908) \\
 & \textbf{SAFDNet (Ours)} & \textbf{0.973 (0.972--0.974)} & \textbf{0.952 (0.950--0.953)} & \textbf{0.934 (0.933--0.936)} & \textbf{0.923 (0.921--0.925)} \\ \bottomrule[2pt]
\end{tabular}%
}
\label{tab:auroc-comparison}
\end{table*}

\subsection{Implementation and Evaluation}
We implemented the proposed SAFDNet model on one NVIDIA Quadro RTX 8000 48G GPU, utilizing the PyTorch framework. The model is selected based on the best checkpoint from the development set. Details for hyperparameters could be found in Appendix. For evaluating the performance of our proposed model, we employ five commonly used metrics: Area Under the ROC curve (AUROC), Area Under the PR curve (AUPRC), Calibration Curve, Accuracy, and F1-score. All metrics were reported with point estimates and 95\% confidence intervals. 
\subsection{Performance Benchmarks}
To better evaluate the performance of our proposed SAFDNet on IOH events early warning, we chose the below algorithms as performance benchmarks in our experiments. For the benchmark models that also for IOH events prediction including MC-CNN~\cite{lee2021deep}, RFM-IBP~\cite{lee2022intraoperative}, CMA~\cite{lu2023composite}, Herein~\cite{hwang2023intraoperative}, TFT~\cite{kapral2024development}, and CNN-Transformer~\cite{yang2024dynamic}, we thus adopted their best results for a fair comparison. We also indicated those missing results as -. We additionally compared our model with various traditional machine learning models and recent deep learning approaches. Specifically, we included classic deep learning models like LSTM and VAE, alongside the GAN-based USAD model~\cite{audibert2020usad} and Transformer-based anomaly detection models such as Anomaly Transformer~\cite{xu2021anomaly} and MEMTO~\cite{song2023memto}. Additionally, we evaluated GDN~\cite{deng2021graph} and Informer~\cite{zhou2021informer}, which are tailored for time series forecasting tasks. We grouped the models for comparison into three subgroups: conventional machine learning models, reconstruction error-based models, and classification-based models. We reproduce these models on VitalDB dataset for a fair comparison with our proposed SAFDNet.

\subsection{Main Results}

The proposed SAFDNet model demonstrates state-of-the-art performance in predicting intraoperative hypotension (IOH) across both the \textit{Only ABP} and \textit{Multi-Channel} settings. SAFDNet consistently outperforms existing benchmarks across all prediction horizons (3, 5, 10, and 15 minutes), underscoring its ability to handle both single-channel and multi-channel biosignal data with high accuracy and robustness.

\subsubsection*{Prediction Performance}

\paragraph{\textit{Only ABP} Setting}  
In the \textit{Only ABP} setting, SAFDNet achieves AUROC values of \textbf{0.968}, \textbf{0.949}, \textbf{0.928}, and \textbf{0.915} for the 3-, 5-, 10-, and 15-minute prediction horizons, respectively (Table~\ref{tab:auroc-comparison}). These results surpass all baseline models, including reconstruction-based methods such as USAD (\textbf{0.924} at 5 minutes) and classification-based methods such as GDN (\textbf{0.929} at 5 minutes). SAFDNet delivers a \textbf{2.5\%} improvement over USAD and a \textbf{2.0\%} improvement over GDN in AUROC at the 5-minute horizon. 

The consistently superior performance of SAFDNet highlights its effectiveness in processing single-channel ABP data. The \textit{Self-Adaptive Filter Block} and \textit{Dual-Path Interactive Attention Block} play pivotal roles in enhancing feature extraction, noise suppression, and temporal modeling, enabling SAFDNet to outperform conventional and deep learning-based approaches.

\paragraph{\textit{Multi-Channel} Setting}  
By incorporating additional biosignals (ABP, ECG, PPG, and CO\textsubscript{2}), SAFDNet achieves even higher performance, with AUROC values of \textbf{0.973}, \textbf{0.952}, \textbf{0.934}, and \textbf{0.923} for the 3-, 5-, 10-, and 15-minute horizons, respectively (Table~\ref{tab:auroc-comparison}). At the 5-minute horizon, SAFDNet outperforms USAD (\textbf{0.935}) by \textbf{1.7\%} and GDN (\textbf{0.934}) by \textbf{1.8\%}. 

The inclusion of multi-channel data allows SAFDNet to leverage complementary information across biosignals, further enhancing its predictive accuracy. This capability underscores its robustness and generalizability, particularly in real-world clinical settings where multi-channel data is often available.
\begin{figure}[!t]
    \includegraphics[width=1\linewidth]{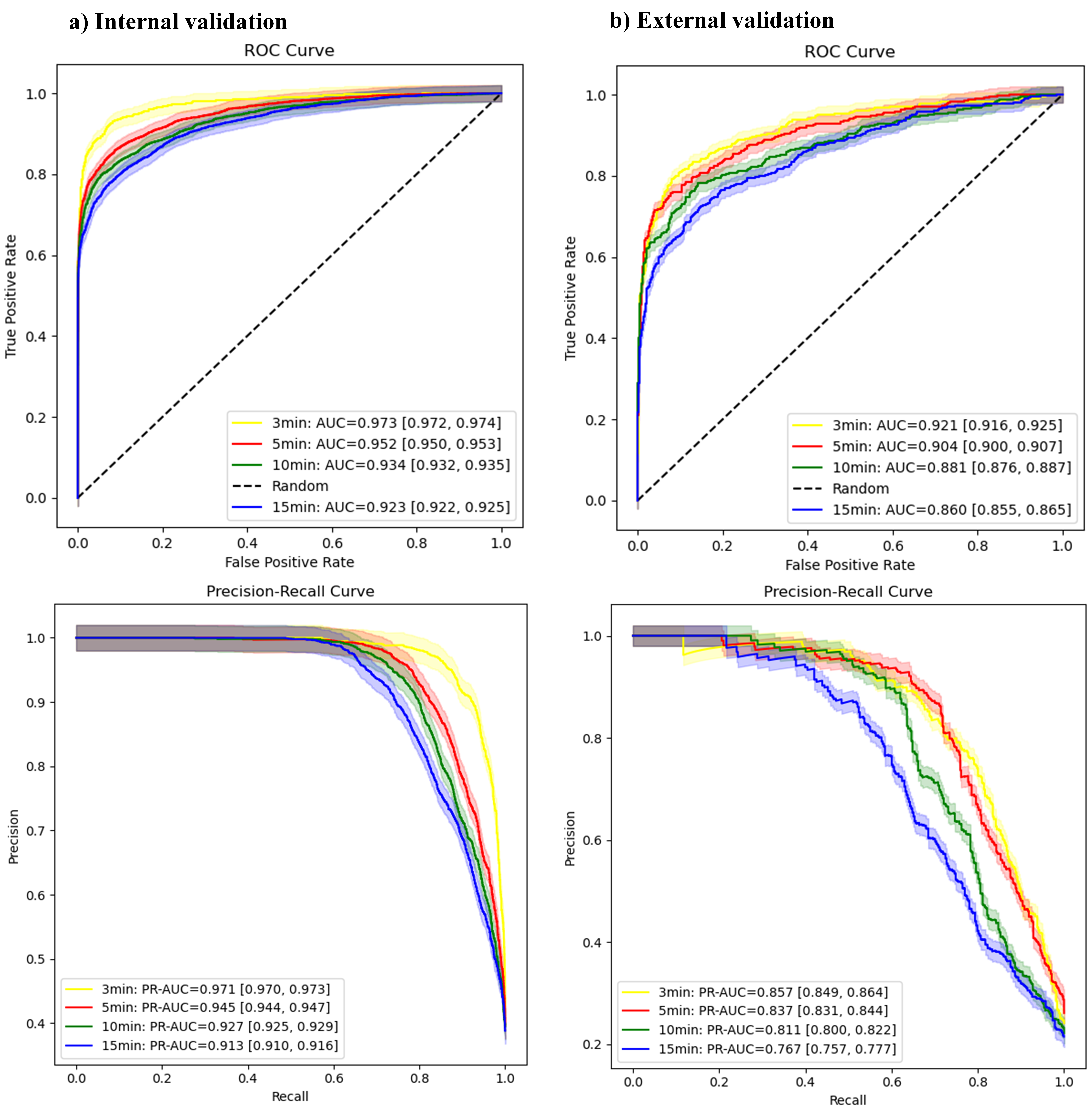}
    \caption{ROC and PR curves in internal validation and external validation.}
    \label{fig:auroc}
\end{figure}
\subsubsection*{Comparison with Public Works}
When compared with publicly available benchmark models for IOH prediction (Table~\ref{tab:performance-comparison}), SAFDNet demonstrates superior performance across all prediction horizons. For example, at the 5-minute horizon, SAFDNet achieves an AUROC of \textbf{0.952}, outperforming MC-CNN~\citep{lee2021deep} (\textbf{0.932}), CMA~\citep{lu2023composite} (\textbf{0.930}), and CNN-Transformer~\citep{yang2024dynamic} (\textbf{0.943}). Moreover, SAFDNet consistently maintains high precision, recall, and F1-scores, indicating its reliability in accurately identifying IOH events while minimizing false positives.

\subsubsection*{Internal and External Validation}  
Figures~\ref{fig:auroc} and~\ref{fig:cali} illustrate SAFDNet’s strong predictive performance in both internal and external validations. 

\textbf{Internal Validation:} SAFDNet achieved AUROC values of \textbf{0.973}, \textbf{0.952}, \textbf{0.934}, and \textbf{0.923} for the 3-, 5-, 10-, and 15-minute horizons, respectively. Corresponding PR-AUC values of \textbf{0.971}, \textbf{0.945}, \textbf{0.927}, and \textbf{0.913} further highlight its precision and reliability.

\textbf{External Validation:} SAFDNet demonstrated strong generalization, achieving AUROC values of \textbf{0.921}, \textbf{0.904}, \textbf{0.881}, and \textbf{0.860} for the 3-, 5-, 10-, and 15-minute horizons, respectively. PR-AUC values, such as \textbf{0.837} for the 5-minute horizon, confirm its robustness in unseen, real-world data.

\subsubsection*{Calibration Analysis}  
\begin{figure}[!t]
    \includegraphics[width=1\linewidth]{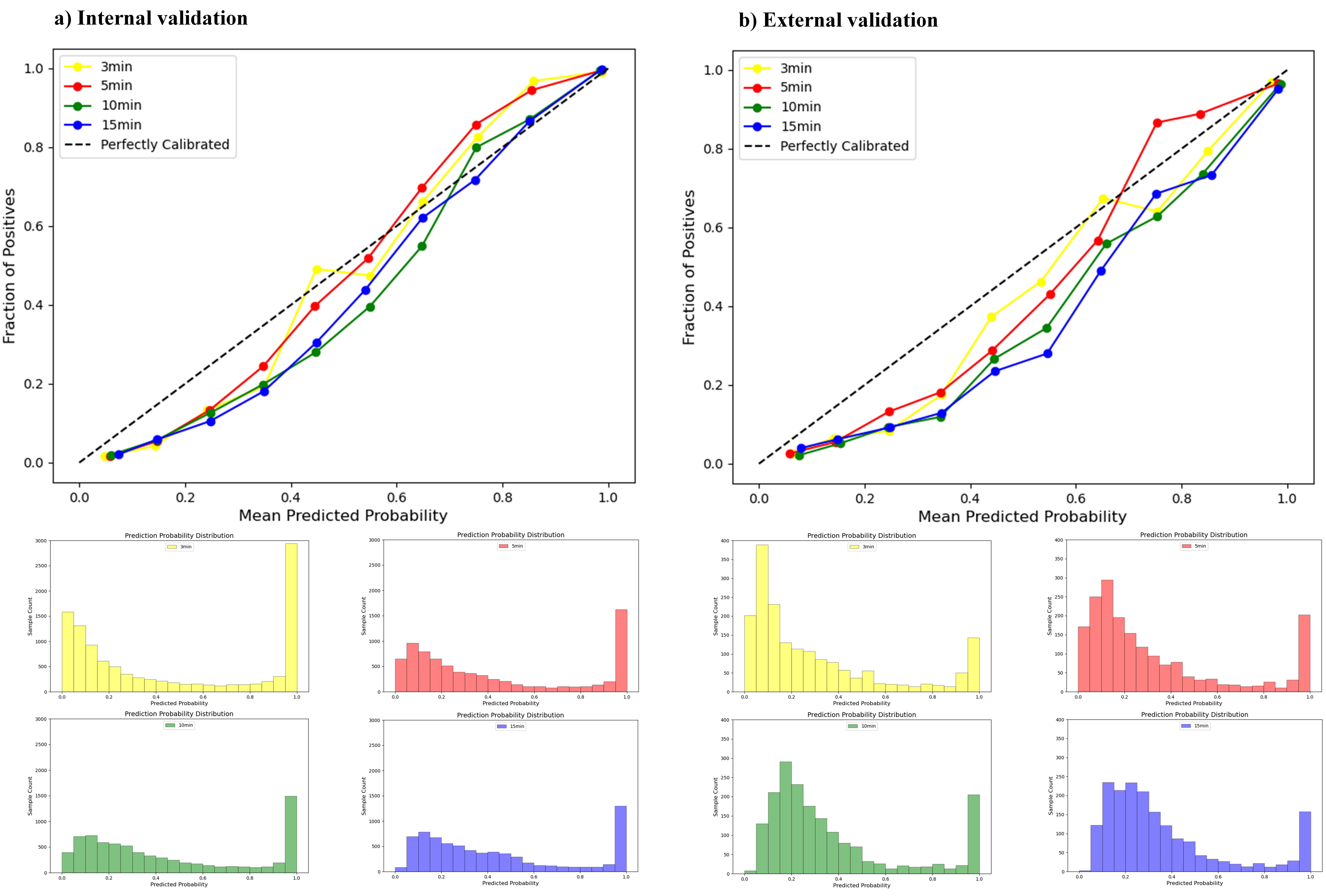}
    \caption{Calibration curves in internal validation and external validation.}
    \label{fig:cali}
\end{figure}

Calibration curves (Figure~\ref{fig:cali}) indicate a strong alignment between predicted probabilities and observed outcomes across all horizons in the internal validation. The corresponding histograms below the calibration curves show the distribution of predicted probabilities at each time interval. The closer the calibration curve is to the dotted line, the better the calibration of the model. The histograms give an indication of the frequency and confidence of the classifier’s predictions. Although the SAFDNet was reasonably calibrated in the internal validation, it overestimated the occurrence of
hypotension in the external validation(Figure~\ref{fig:cali}). Miscalibration is a common phenomenon when predictive
models are tested in a population that they were not developed in, highlighting that predictive models should be carefully tested prior to implementation into clinical practise. This overestimation in the external validation is not necessarily an error of the SAFDNet but is rather a reflection that the model is not anticipating future medical interventions, even though we trained the model on retrospective surgical data. Before the deployment of the SAFDNet in real clinical application, we should use recalibration method. After recalibration, we believe the SAFDNet can generate well-calibrated predictions, ensuring confidence in its utility for clinical decision-making. The model’s ability to consistently separate IOH and non-IOH cases with high confidence further validates its effectiveness as a practical tool for early IOH prediction.

\begin{table}[!t]
\centering
\caption{Performance comparison of our method with the benchmark models that also for IOH events prediction.}
\resizebox{\linewidth}{!}{%
\begin{tabular}{c|c|c|c|c|c|c}
\hline
\textbf{Method} & \textbf{Time} & \textbf{AUROC} & \textbf{Accuracy} & \textbf{Precision} & \textbf{Recall} & \textbf{F1-score} \\ \hline
\multirow{4}{*}{\textbf{SAFDNet (Ours)}} 
 & 3min  & \textbf{0.973} & \textbf{0.917} & \textbf{0.915} & \textbf{0.892} & \textbf{0.903} \\ 
 & 5min  & \textbf{0.952} & 0.893 & \textbf{0.907} & \textbf{0.814} & \textbf{0.858} \\ 
 & 10min & \textbf{0.934}          & \textbf{0.873}          & \textbf{0.850}         & 0.818          & \textbf{0.834}          \\ 
 & 15min & \textbf{0.923}          & \textbf{0.862}          & \textbf{0.837}          & 0.787          & \textbf{0.811}          \\ \hline
\multirow{3}{*}{\textbf{MC-CNN~\citep{lee2021deep}}} 
 & 5min  & 0.932          & -              & 0.858          & \textbf{0.858}          & -              \\ 
 & 10min & 0.912          & -              & 0.834          & \textbf{0.835}          & -              \\ 
 & 15min & 0.897          & -              & 0.814          & \textbf{0.814}          & -              \\ \hline
\textbf{RFM-IBP~\citep{lee2022intraoperative}} 
 & 5min  & -              & \textbf{0.989}         & 0.652          & 0.441          & -              \\ \hline
\multirow{3}{*}{\textbf{CMA~\citep{lu2023composite}}} 
 & 5min  & 0.930          & 0.906          & 0.655          & 0.750          & -              \\ 
 & 10min & 0.858          & 0.861          & 0.509          & 0.605          & -              \\ 
 & 15min & 0.834          & 0.860          & 0.500          & 0.525          & -              \\ \hline
\textbf{Herein~\citep{hwang2023intraoperative}} 
 & 10min & 0.904          & -              & -              & -              & -              \\ \hline
\multirow{4}{*}{\textbf{TFT~\citep{kapral2024development}}} 
 & 1min  & 0.960 & -              & -              & -              & -              \\ 
 & 3min  & 0.945          & -              & -              & -              & -              \\ 
 & 5min  & 0.903          & -              & -              & -              & -              \\ 
 & 7min  & 0.867          & -              & -              & -              & -              \\ \hline
\multirow{3}{*}{\textbf{CNN-Transformer~\citep{yang2024dynamic}}} 
 & 5min  & 0.943          & -              & -              & -              & -              \\ 
 & 10min & 0.928          & -              & -              & -              & -              \\ 
 & 15min & \textbf{0.923}          & -              & -              & -              & -              \\ \hline
\end{tabular}%
}
\label{tab:performance-comparison}
\end{table}

\section{Discussion}
\subsection{Effect of Self-Adaptive Filter Block}
To assess the role of the Self-Adaptive Filter Block (SAFB) in SAFDNet, we conducted Ablation Experiment 1, which removed SAFB to disregard frequency-domain features. As shown in Table~\ref{tab:unified-ablation}, the absence of SAFB led to a significant performance decline across both datasets, highlighting its critical contribution. 

On the internal VitalDB dataset, removing SAFB reduced AUROC by 0.7\%–1.2\%, while on the external ZhongdaVital dataset, the drop was more pronounced at 1.2\%–1.8\%. For example, at the 5-minute prediction horizon, AUROC decreased from \textbf{0.952} to \textbf{0.945} on VitalDB and from \textbf{0.904} to \textbf{0.892} on ZhongdaVital. F1 scores also declined consistently, demonstrating SAFB’s ability to improve the precision and reliability of IOH predictions.

\begin{figure}[!t]
    \centering
    \includegraphics[width=0.5\textwidth]{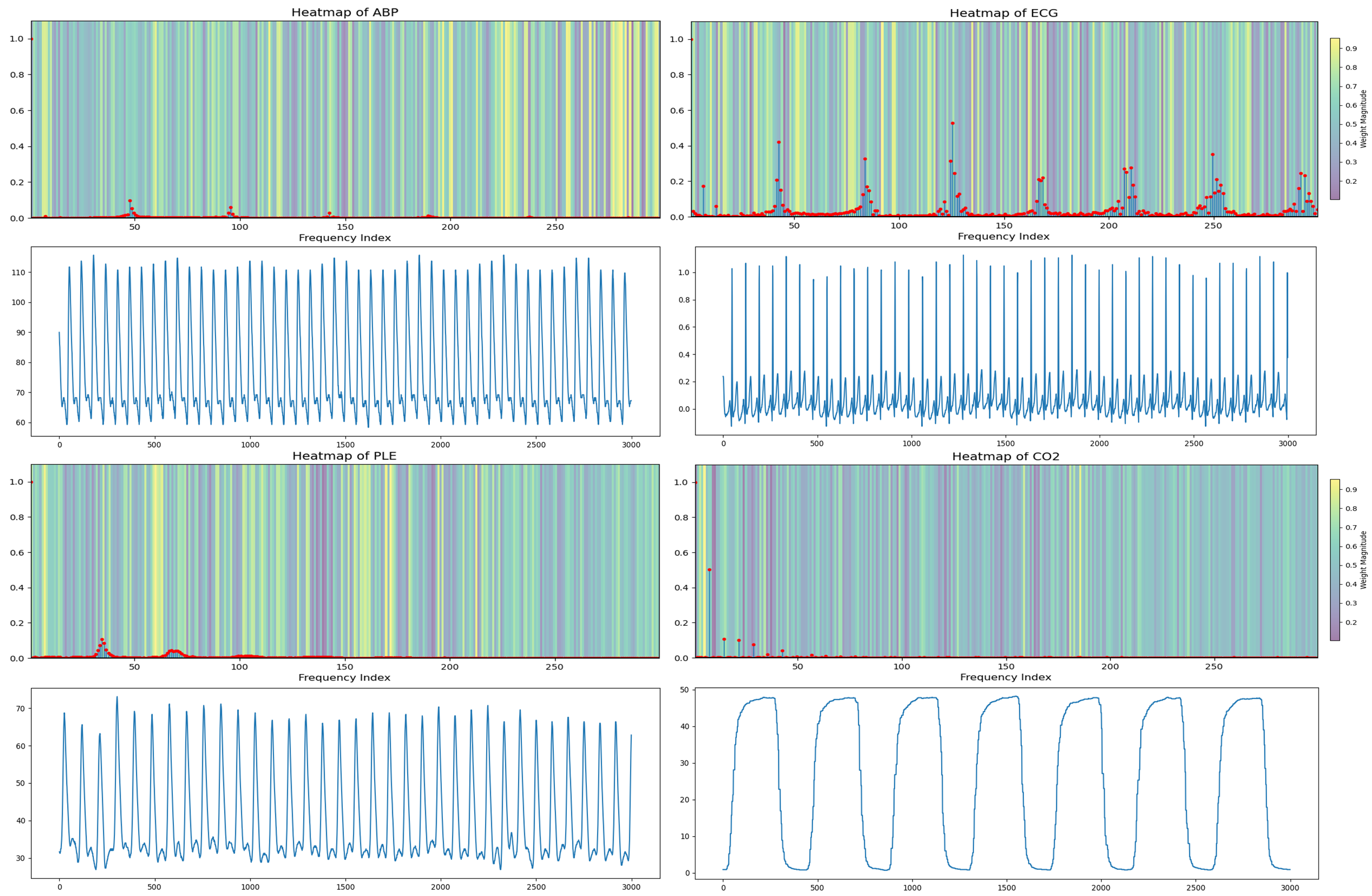}
    \caption{Visualization of ABP, ECG, PPG and CO2 in frequency domain, where the gradual transition of colors which from yellow to purple reflects a continuous increase in the coefficient values.}
    \label{fig:vis}
\end{figure}

The SAFB effectively suppresses noise while extracting critical frequency-domain features. It learns individualized thresholds for biosignals such as ABP, ECG, PPG, and CO2, enabling the model to filter irrelevant noise and emphasize subtle frequency patterns linked to IOH events. For instance, in ABP waveforms, SAFB identifies abnormal frequency bands associated with impending hypotension, as shown in Figure~\ref{fig:vis}. This capability is crucial in clinical settings where biosignals are often noisy and variable.

The larger performance degradation on the external ZhongdaVital dataset underlines SAFB’s robustness in handling real-world noise and variability. By enabling SAFDNet to integrate frequency-domain information and mitigate noise, SAFB ensures consistent and superior predictive performance across diverse datasets. These results affirm SAFB as a vital component of SAFDNet and emphasize its practical value in achieving reliable IOH predictions, even in challenging clinical environments.

\begin{table}[!t]
\centering
\caption{Ablation results for VitalDB and ZhongdaVital datasets.}
\resizebox{\linewidth}{!}{%
\begin{tabular}{c|c|ccc|ccc}
\hline
\multirow{2}{*}{\textbf{Method}} & \multirow{2}{*}{\textbf{Time}} & \multicolumn{3}{c}{\textbf{VitalDB}} & \multicolumn{3}{c}{\textbf{ZhongdaVital}} \\ \cline{3-8} 
 &  & \textbf{AUROC} & \textbf{Accuracy} & \textbf{F1} & \textbf{AUROC} & \textbf{Accuracy} & \textbf{F1} \\ \hline
\multirow{4}{*}{\textbf{SAFDNet}} & 3min & \textbf{0.973} & \textbf{0.917} & \textbf{0.903} & \textbf{0.920} & \textbf{0.896} & \textbf{0.769} \\ 
 & 5min & \textbf{0.952} & \textbf{0.893} & \textbf{0.858} & \textbf{0.904} & \textbf{0.885} & \textbf{0.751} \\ 
 & 10min & \textbf{0.934} & 0.873 & \textbf{0.834} & \textbf{0.881} & \textbf{0.877} & \textbf{0.717} \\ 
 & 15min & 0.923 & 0.862 & \textbf{0.811} & \textbf{0.856} & 0.862 & \textbf{0.679} \\ \hline
\multirow{4}{*}{\textbf{Ablation1}}& 3min & 0.961 & 0.899 & 0.877 & 0.911 & 0.890 & 0.752 \\  
 & 5min & 0.945 & 0.884 & 0.849 & 0.892 & 0.880 & 0.723 \\ 
 & 10min & 0.927 & 0.871 & 0.820 & 0.869 & 0.863 & 0.696 \\ 
 & 15min & 0.918 & 0.860 & 0.799 & 0.838 & \textbf{0.869} & 0.667 \\ \hline
\multirow{4}{*}{\textbf{Ablation2}} & 3min & 0.963 & 0.901 & 0.878 & 0.912 & 0.889 & 0.758 \\
 & 5min & 0.947 & \textbf{0.893} & 0.853 & 0.896 & 0.873 & 0.724 \\ 
 & 10min & 0.927 & \textbf{0.877} & 0.826 & 0.880 & 0.870 & 0.707 \\ 
 & 15min & 0.913 & \textbf{0.866} & 0.804 & 0.847 & 0.852 & 0.661 \\ \hline
\multirow{4}{*}{\textbf{Ablation3}} & 3min & 0.966 & 0.898 & 0.891 & 0.909 & 0.869 & 0.762 \\
 & 5min & 0.948 & 0.890 & 0.851 & 0.891 & 0.859 & 0.723 \\ 
 & 10min & 0.928 & 0.876 & 0.828 & 0.874 & 0.842 & 0.692 \\ 
 & 15min & 0.920 & 0.863 & 0.807 & 0.839 & 0.837 & 0.657 \\ \hline
\end{tabular}%
}
\label{tab:unified-ablation}
\end{table}
\subsection{Effect of Dual-Path Interactive Attention Block}
The impact of the Dual-Path Interactive Attention Block (DPIAB) was evaluated through Ablation Experiments 2 and 3, which removed its components from SAFDNet. As shown in Table~\ref{tab:unified-ablation}, the absence of DPIAB resulted in a notable performance drop across both datasets. AUROC declined by 0.5\%–1.0\% on VitalDB and 0.8\%–1.5\% on ZhongdaVital. For instance, at the 10-minute horizon, AUROC decreased from \textbf{0.934} to \textbf{0.928} on VitalDB and from \textbf{0.881} to \textbf{0.874} on ZhongdaVital.

DPIAB enables SAFDNet to capture both short-term and long-term temporal dependencies using a dual-path design that combines CNN-based local pattern extraction with LSTM-based global sequence modeling. This synergy improves temporal feature representation, resulting in higher accuracy and robustness.

The larger performance drop on ZhongdaVital highlights DPIAB’s importance in handling real-world clinical variability. By integrating DPIAB, SAFDNet achieves superior predictive performance, making it well-suited for practical perioperative applications.

\subsection{Clinical Applicability of the Proposed SAFDNet}
The experimental results from internal (VitalDB) and external (ZhongdaVital) validations demonstrate that SAFDNet maintains consistent, high predictive performance across diverse datasets, showcasing its strong generalization capability. This robustness is crucial for real-world clinical scenarios, where data variability and noise are common challenges.

\begin{figure}[!t]
\centering
    \includegraphics[width=\linewidth]{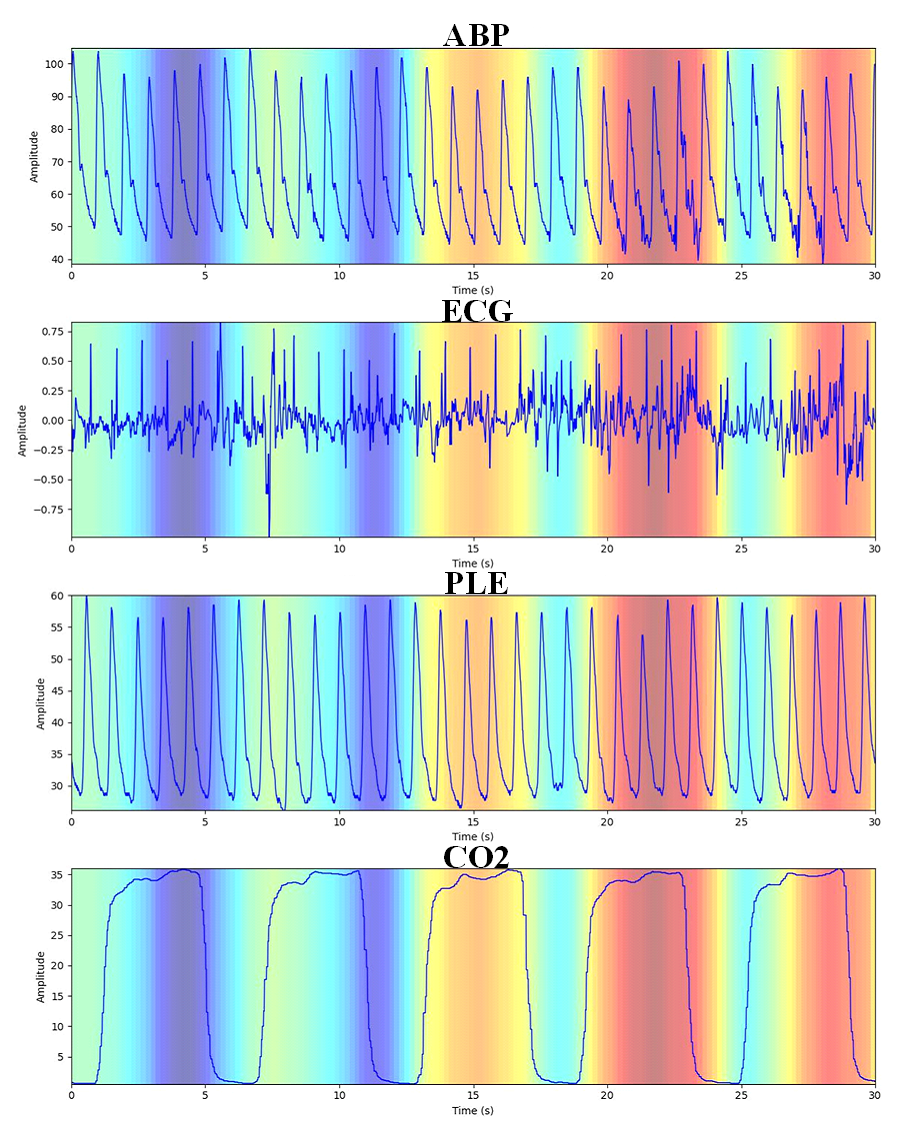}
    \caption{Sensitivity map for ABP, ECG, PPG and CO2, where the gradual transition of colors which from blue to yellow to red reflects a continuous increase in the coefficient values.}
    \label{fig:grad-cam}
\end{figure}

In addition to its predictive performance, SAFDNet provides interpretability, a key factor for clinical adoption. The adaptive filter weights (Figure~\ref{fig:vis}) highlight how SAFDNet suppresses noise while identifying critical frequency bands associated with IOH. Furthermore, sensitivity maps generated using Grad-CAM (Figure~\ref{fig:grad-cam}) reveal the temporal regions most relevant to IOH prediction, offering a deeper understanding of the model's decision-making process. These visualizations enhance clinician trust in SAFDNet by addressing the "black-box" nature of deep learning models.

Overall, the combination of robust prediction, generalizability, and interpretability indicates that SAFDNet is well-suited for deployment in real-world perioperative settings. By providing accurate early warnings for IOH, SAFDNet has the potential to support proactive clinical interventions and improve patient outcomes.

\section{Conclusion}
Intraoperative hypotension (IOH) poses significant risks to perioperative patients, necessitating timely and accurate prediction to enable proactive clinical interventions. In this study, we proposed \textbf{SAFDNet}, a novel framework designed to address key limitations of existing IOH prediction methods, including underutilization of frequency-domain information, suboptimal temporal dependency modeling, and sensitivity to noise.

SAFDNet integrates a \textbf{Self-Adaptive Filter Block}, which leverages Fourier analysis to extract frequency-domain features while suppressing noise, and a \textbf{Dual-Path Interactive Attention Block}, which combines CNN-based local pattern extraction with LSTM-based global dependency modeling. This architecture enables SAFDNet to effectively predict IOH events across diverse scenarios and datasets.

Extensive experiments on two large-scale real-world datasets (VitalDB and ZhongdaVital) demonstrated SAFDNet’s superior performance, achieving AUROC values of up to \textbf{97.3\%} for IOH early warning, consistently outperforming state-of-the-art models. SAFDNet also exhibited strong generalization capabilities in external validation, robust noise resilience, and high interpretability through adaptive filtering and sensitivity maps.

These results highlight SAFDNet’s potential for clinical deployment, offering a reliable and actionable solution for IOH early warning. By providing accurate predictions and enhancing interpretability, SAFDNet supports a shift from reactive to proactive perioperative care, ultimately improving patient outcomes.
\appendix

\section*{Ethical Statement}

The internal validation dataset used in this study, VitalDB, is publicly available and was accessed through PhysioNet after completing the required ethical training program. As such, this work adheres to ethical guidelines, and we do not foresee any ethical concerns arising from its use. The external validation dataset ZhongdaVital was approved by the Institutional Review Board of Zhongda Hospital affiliated with Southeast University (No. 2024ZDSYLL476-P01), and written informed consent was waived due to minimal risk to participants.



\begin{ack}
This work is supported by Provincial Natural Science Foundation of Jiangsu (NO. BK20240703) , and the Startup Foundation for Introducing Talent of NUIST (NO. 1523142401057, 1523142401055). The code and supplementary material are available at https://github.com/sleepwithwind/safdnet. 
\end{ack}



\bibliography{mybibfile}

\end{document}